\documentclass[a4paper, 11pt]{report}
\usepackage{setspace}
\usepackage{fancyhdr}
\usepackage{hyperref}
\usepackage{graphicx}
\author{\textbf{Alexander Darer}\\* Supervised by: Dr Peter Lewis}
\title{Mini Project 1:\linebreak \textbf{A Cellular Automaton Based Controller for a Ms. Pac-Man Agent}}

\begin{document}

\maketitle

\begin{abstract}
Video games can be used as an excellent test bed for Artificial Intelligence (AI) techniques. They are challenging and non-deterministic, this makes it very difficult to write strong AI players. An example of such a video game is Ms. Pac-Man.

In this paper I will outline some of the previous techniques used to build AI controllers for Ms. Pac-Man as well as presenting a new and novel solution. My technique utilises a Cellular Automaton (CA) to build a representation of the environment at each time step of the game. The basis of the representation is a 2-D grid of cells. Each cell has a state and a set of rules which determine whether or not that cell will dominate (i.e. pass its state value onto) adjacent cells at each update. Once a certain number of update iterations have been completed, the CA represents the state of the environment in the game, the goals and the dangers. At this point, Ms. Pac-Man will decide her next move based only on her adjacent cells, that is to say, she has no knowledge of the state of the environment as a whole, she will simply follow the strongest path.

This technique shows promise and allows the controller to achieve high scores in a live game, the behaviour it exhibits is interesting and complex. Moreover, this behaviour is produced by using very simple rules which are applied many times to each cell in the grid. Simple local interactions with complex global results are truly achieved.
\end{abstract}

\thispagestyle{empty}
\section*{Acknowledgements}
I would like to thank my supervisor - Dr Peter Lewis, whose knowledge, guidance and experience has been invaluable throughout this project.

\tableofcontents

\listoffigures


\chapter{Introduction to the Project}

Artificial Intelligence (AI) is a huge field of Computer Science. It ranges from search to machine learning, robotics to networking and of course, video gaming. Video gaming has become a huge market in a fairly short space of time as game developers endeavour to produce original and enjoyable games year on year. One way they can achieve this is by building a 'smart' AI player which can challenge any person it plays against, moreover, the enjoyment of a game can be increased if an AI player can achieve 'human-like' game-play (i.e. human players and AI players are indistinguishable). This has proved to be a very difficult problem, building strong AI players which can react positively to other players and the game itself has proven to be incredibly complex. It seems at this present time that even the strongest AI players (when AI cheating is disallowed) are massively outperformed by the best human players. There is therefore a need for new and improved AI techniques.

\section{Novel Approaches to Video Game AI}
This project will introduce some previous research and approaches to game AI (specific examples which apply to Ms. Pac-Man) and will describe a novel technique which has not generally been applied to video game AI before. The technique will be tested in a live game situation and the behaviour of the player will be examined for interesting and unusual traits. The idea behind this research is to learn whether or not certain novel AI techniques, which have not been used widely in video games, can be as effective as traditional algorithms. There will be discussions about different examples of AI which have been used to build Ms. Pac-Man controllers, some of which are more traditional in video game AI and some which have not been widely used in this context. 

\section{Proposed Questions and Contributions}

In this project I will try to show why it is necessary to spend time researching the benefits of applying novel techniques to game AI. I will contrast and compare my approach to others in the same context and will hopefully show that it has promise for future research. This notion of applying a novel technique to game AI will contribute a fresh perspective for the field, akin to what other researchers have started undertaking.\linebreak

{
\raggedleft
\textbf{Formally, the questions I hope to answer are:}
}

\begin{itemize}
\item Why apply novel techniques to game AI?
\item Why use Ms. Pac-Man as a test bed for AI?
\item How successful was a Cellular Automaton based approach to the problem?
\end{itemize}

\subsection{What Knowledge will I Gain from this Research?}

During this project I will develop several skills which are required to complete the research. Firstly, I will need to critically analyse research papers so I have an understanding of the current research which is being completed in this field. I will need to learn what different techniques are being attempted, their advantages and disadvantages, so I can decide where I will execute my own research into a novel technique. I will also need to spend time learning the API for the simulator which will be used to run and test my approach (since this has been developed by a third party). Secondly, I will improve my programming skills as my AI technique will be written from the ground up and will also need to implement a software interface so it can integrated into the simulator program. Finally, I will analyse my own approach, I will run a number of different experiments to determine its performance and will then examine its behaviour to conclude if it shows promise for future research. All of this research will be published in this report which will strengthen my writing skills. 

\section{Current Conferences and Research}

There is a lot of active work in this field being currently undertaken. Many researchers are building AI controllers for Ms. Pac-Man and submitting them to various different competitions (some of which are run at conferences). The conferences which are currently most active are the IEEE Conference on Computational Intelligence and Games and the IEEE Congress on Evolutionary Computation. There have been competitions held at both of these conferences. Overall, many approaches have yielded good results, some very strong AI players have been produced using various different techniques. They are still however, fairly weak when compared to the best human players.

\section{How will the Approach will be Evaluated?}

My approach will be used to power a Ms. Pac-Man agent in live game situations. There will be many different forms of evaluation its performance. Firstly there will be the scores it achieves. This will allow us to compare it against other forms of Ms. Pac-Man controllers in a simply score comparison. Secondly, the behaviour it exhibits will be explained and analysed for desirable traits which could show promise. On the other hand, any undesirable behavioural traits will discussed. There will be different control cases which will be used to for comparison for both scores and behaviour, these will be explained in the later sections.




\chapter{Literary Review}

Game AI has been a very active field of research for many, many years now. A huge amount of time and money has been invested into developing 'smart' AI players for a wide genre of games. Aside from the obvious perks of strong game AI (as in more enjoyment for human players when the AI gives them a challenge), the techniques developed can be applied to other problems in computer science. Video games which are non-deterministic and require adaptation, resilience and knowledge to beat (or even play) can be used as very effective test beds when developing new AI techniques or improving older ones. Simple games such as \emph{Ms. Pac-Man}\textsuperscript{\cite{mspman}} and \emph{Super Mario}\textsuperscript{\cite{mario}} or more complex games such as \emph{TORCS (The Open Racing Car Simulator)}\textsuperscript{\cite{torcs}}, \emph{Unreal Tournament}\textsuperscript{\cite{unre}} and \emph{Starcraft}\textsuperscript{\cite{star}} can all be used for the purpose of developing, improving and testing AI.

In recent years the simple video games have been put to widespread use as test beds for researching many forms of AI. Ms. Pac-Man has been a very popular choice among researchers because it is simple to play but very hard to be good at, it has been proven to be extremely difficult to build strong AI players (compared to the best human players). This is due to the large number of different problem characteristic associated with Ms. Pac-Man. To benchmark the different approaches which have been developed, tournaments are held annually at various different conferences, notably the IEEE Conference on Computational Intelligence and Games (CIG) and the IEEE Congress on Evolutionary Computation (CEC). This gives researchers a chance not only to see other approaches in action, but to compete against them with their own AI players. It is at these conferences where you will find some of the latest, cutting-edge research in this field.



\section{The Game}

\emph{Pac-Man}\textsuperscript{\cite{pman}} is a single player video game developed by Namco in the 1980's. It was originally released on the Namco Pac-Man Arcade System. Since then Pac-Man has become one of the most popular and well known video games ever released, a true arcade classic. Ms. Pac-Man is a variation of the original game, released in 1982 it has the same rules and goals as the original Pac-Man. The only difference between Pac-Man and Ms. Pac-Man (except the bow) is that the ghosts in Ms. Pac-Man are non-deterministic as opposed to the deterministic ghosts in Pac-Man. This is a very important distinction because a player could simply remember the route the ghosts would take and thus beat the original game. By introducing non-determinism, Ms. Pac-Man becomes a lot harder, players need to be adaptive and reactive to the ghosts. A player cannot predict what the state the game will be in at any given time step and thus cannot guess the movements of the ghosts.

\subsection{Rules of Ms. Pac-Man}

Ms. Pac-Man is a very simple game to play. You (as the player) have to move the Pac-Man (yellow figure) around the maze and collect all of the Pills (small white dots) and Power-Pills (large white dots) whilst avoiding the ghosts (red, white, blue and pink figures) at all times. Once all of the pills have been collected you would move onto the next level or maze, however if at any point you come into contact with a ghost, you will loose a life. If you loose all of your lives the game will end. When you consume a Pill your score will increase, when a Power-Pill is consumed the ghosts will turn edible allowing you to consume them and score extra points, once a ghost is consumed it will return to the ghost lair and will no-longer be edible (until another Power-Pill is consumed).

\subsection{Ms. Pac-Man vs Ghosts}

The \emph{Ms. Pac-Man vs Ghosts League}\textsuperscript{\cite{pvgl}} is a competition which is held at the conferences explained earlier. The researchers behind the competition provide a simulator for Ms. Pac-Man so participants only need to code their AI controllers (for either Pac-Man or the ghosts). The simulator provides all of the other functionality behind the game including the game engine, the game state, visual effects (no sound unfortunately), etc. At each time step of a running game the controllers are handed a copy of the current game state and they will return a move which will then be executed by the game engine. The program is written in Java.

This simulator is a fairly recent addition for the competition which used to use a screen-capture method of extracting data about the game state. In this method a game of Ms. Pac-Man was ran using the competitors AI controller which had to get its entire input about the game state from a screen-shot of the game window. This is inherently less efficient than the new simulator as a great deal of image processing had to be completed before the controller could even consider its next move.



\section{Cellular Automata}

Cellular Automata (CA's) are nature inspired systems which can produce incredibly complex behaviour from simple, localised interactions between cells. Cells are constituents of a larger system - a lattice of cells (or array), each cell contains a discrete value which can be manipulated using a set of rules. The rules usually take into account the neighbours of the cell. Stephen Wolfram explains that Cellular Automata are "mathematical idealisations of physical systems in which space and time are discrete, and physical quantities take on a finite set of discrete values".\cite{step}

A CA can be used to model many different systems in any number of dimensions - though usually only 1-D, 2-D or 3-D. An update on CA means that the cells in the CA have their values updated according to a set of rules. The values a cell can take on and the rules associated with it depend completely on the application or implementation. What is true however, is that any CA can produce globally complex, self-organising behaviour which is caused only by the simple interactions between individual cells. 

\begin{figure}[h!]
  \centering
    \includegraphics[width=0.5\textwidth]{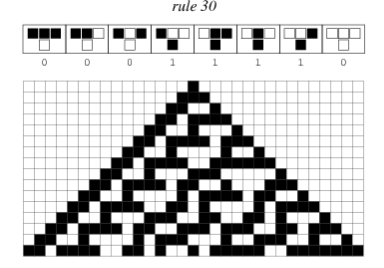}
  \caption{\small{A simple 1-D cellular automaton. Taken from Wolfram Mathworld\cite{wolf}.}}
  \label{fig:c1}
\end{figure}

In Figure~\ref{fig:c1} there is a simple 1-D CA which shows how a simple rule can produce something much more complicated. Each horizontal in the image shows one update step on the CA.

\begin{figure}[h!]
  \centering
    \includegraphics[width=0.5\textwidth]{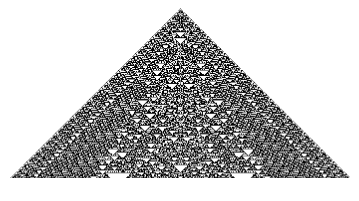}
  \caption{\small{A more complicated 1-D cellular automaton. Taken from Wolfram Mathworld\cite{wolf}}}
  \label{fig:c2}
\end{figure}

Figure~\ref{fig:c2} shows a CA with simple rules, similar to that in Figure~\ref{fig:c1}, however this image shows many more updates on the CA. It produces a hugely complex lattice of cells which results only from simple rules and simple, local interactions.

\section{Previous Approaches}
 
 
There have been numerous different approaches to design Ms. Pac-Man agents, some have been very successful, some not so much. The different techniques which have been employed are varied, from nature inspired Evolutionary Computation to Agents and Rule-Based approaches. There has indeed been a large amount of research focused on experimenting with wildly different approaches, each resulting in peculiar but interesting conclusions.

 
\subsection{Rule-Based Approaches}
A common approach to building a successful agent is adopting rule-based policies. This is a relatively simple solution, although the rules which are used can be very expansive and complex. The rules can be either hand-coded or generated through a computer program (e.g. a genetic algorithm). The complexity of the rules depends on the specific goals which want to be achieved. Building rule-based approaches brings up another problem however; deciding the parameters which the rules use. The process of optimising parameters which the rules use is a task which can be just as difficult as building the rules themselves and when you reach a set of optimised parameters, how do you know if they are the global optimum? The process of solving this problem can be addressed in a number of different ways, some of which will be discussed here.

\subsubsection{Hand-Coded Rules}
Hand-coding rules for a game agent can be a very lengthy process, this is due to the fact that you must encode every action the agent will need to make. Even in a relatively simple game like Ms. Pac-Man, there are a lot of different actions and observations which will need to be accounted for and subsequently programmed into your rule set. Having said this, the outcome of designing your own rule based system can produce a large payout if you design clever tactics for your agent. \emph{ICE Pambush 2}\textsuperscript{\cite{pam2}}, designed by Ruck Thawonmas and Hiroshi Matsumoto, won the \emph{IEEE CEC 2009 Software Agent Ms. Pac-Man Competition}\textsuperscript{\cite{pam2}}, their approach used only hand-coded rules and parameters. 

Pambush 2 was written to compete using the older version of the Ms. Pac-Man simulation software, this meant that the authors had to create an Image processing module to 'capture' information about the current state of the game. With the release of the new Ms. Pac-Man simulator, this screen-capture module is no longer necessary as the game state information is available directly from the simulator. The other module within Pambush 2 contains the rules to control the agent and executes decisions based on these rules. 

Pambush 2 uses 7 rules in total, the first rule being the highest priority and the 7th being the lowest. Each rule has a set of conditions and then an action which would be executed if all of the conditions pass. The actions are targets which the agent should move towards. At a first glance the rules look deceptively simple, as they seem to try to guide the agent away from the ghosts and towards edible pills and power pills. However, the authors of Pambush 2 have encoded some very interesting tactics where the agent will attempt to ambush the ghosts by allowing them to get close, eating a power pill and then consuming the edible ghosts. This sort of behaviour allowed the agent to build up large scores by consuming many ghosts over the course of the game. To help the agent decide which rule should be used, there are cost definitions which define how likely it is that the agent will reach its target without being eaten by a ghost. This system was entirely hand-coded and shows the promise of this approach, which may seem simple, but becomes incredibly difficult as the tactics which are to employed become more complex.

The parameters used for the rules in Pambush 2 are also hand-picked and optimised, for example, in the first rule, there is the following condition:\linebreak
\centerline{\texttt{distance(nearest\_ghost) \textgreater = 4}}\linebreak
The given parameter for this rule was hand-picked to try to give the greatest payout, but how do the authors know if this is the most optimum parameter for this condition? This is something they considered in the next generation of their agent, called \emph{ICE Pambush 3}\textsuperscript{\cite{pam3}}.

Pambush 3 uses a hand-coded rule system similar to Pambush 2, however not only have the rules been tweaked and improved, the parameters are now automatically optimised by evolutionary means. Pambush 3 uses a different set of parameters for each level in the game, each set was individually optimised to suit it's respective level. By doing this, the system can play each level to a higher standard than if the parameters are kept constant. The authors report that even for just the first level in the game, the system which used optimised parameters achieved a 17\% performance increase when compared to the same system using manually optimised parameters. This shows that using an automated system to find (or get as close to) the globally optimum parameters is much more efficient and successful  than attempting to find them manually.
  
\subsection{Evolutionary Computation}
There have been many examples in recent years of Ms. Pac-Man agents based on \emph{Evolutionary Computation}\textsuperscript{\cite{gramma, EA1, trainingcamp, inforet}}. Evolutionary Algorithms can be used to evolve individual agents or optimise parameters that control the agents behaviour - like a significant number of the examples discussed here. Previous research has shown the promise of using evolutionary techniques, they can yield successful results and can be applied to a huge range of combinatorial and multi-objective problems. One of which, is Ms. Pac-Man.

\subsubsection{Genetic Programming}
\emph{Genetic programming (GP)}, a branch of Evolutionary Computation, has been widely used in the development of Ms. Pac-Man agents. Since it's introduction \textsuperscript{\cite{GP1}}, GP has been featured in many machine intelligence applications, there has been a huge amount of research and development into the field. One of the first uses of GP as an automated controller for Pac-Man was described by \emph{Koza}, in his first GP book\textsuperscript{\cite{GP2}}. His controllers were relatively basic and the simulator he used was a simplified version of the original game, however, his work can be used a foundation to build more advanced and better performing agents. Atif Alhejali and Simon Lucas, of the Game Intelligence Group at the University of Essex, have developed a diverse and successful controller\textsuperscript{\cite{EA1}}, using GP, based on Koza's original experiment\textsuperscript{\cite{GP2}}.

Alhejali and Lucas's approach uses a function set which was derived from Koza's experiment, it follows a similar idea but has been majorly modified from the original version. Their function set contains different groups: functions, which help the agent make decisions and terminals, which are split into two sub-groups; data-terminals and action terminals. The data-terminals provide information about the current state of the game and action terminals specify an action that Pac-Man will make during the game. There were three different experiments used to produce results: a single maze experiment; a four maze experiment and an unlimited maze experiment (which was a repetition of the maze in the first experiment over and over). During the testing of every experiment the agents were given only one life.

The GP parameters used for the experiment are relatively simple - population size of 1000, cross-over rate of 80\%, mutation rate of 20\% and then run for 50 generations. The selection process used is tournament selection with a tournament size of 3. To calculate the fitness for each individual, the algorithm runs the simulator five times for each generated agent and then the average score is used. The authors comment on how the value of 5 was chosen to try to achieve the best results from the noisiness of evaluation caused by Ms. Pac-Man being very non-deterministic, while not compromising too much CPU time. At the end of the GP run, there is a second fitness evaluation to ensure the best agent is chosen. The individuals will qualify for this second evaluation if their fitness is greater than an evaluator value (a hand picked score). Once qualified, new fitnesses are calculated for the agents based on the average of 100 runs, the highest scoring agent is then declared the best agent from the GP run.

To compare against the generated controllers, the authors produced a hand-coded version of an agent which uses the same function set available to the others. This was to show how an evolved controller would perform in contrast to a hand-written counterpart.

The results Alhejali and Lucas obtained varied quite significantly for each of the different experiments. In the \emph{single-maze} experiment, the agent focused mainly on eating power pills and then hunting the edible ghosts. Once the power pills had been used up many of the agents appeared to become stuck and not 'know' what to do. This was caused by the fact that in the single maze the points that could be scored by eating the ghosts far outweighed the points available by eating pills. The better agents were able to compromise between eating ghosts and eating pills, while staying safe by keeping away from inedible ghosts. In the \emph{four-maze} experiment the behaviour of the best agents was very similar to those from the single-maze experiment. The agents from this experiment however achieved higher average scores overall, this may have been due to the fact that the agents were evolved in different environments (mazes) which allowed them to evolve better traits throughout. Since the agent could progress onto different levels, the behaviour of the agent prioritises staying alive, rather than building up a huge score. In the \emph{unlimited-maze} experiment, agents were produced with completely different behaviour compared to the first two experiments. Since the agents could continue through unlimited mazes, the best strategy was to clear as many as possible, rather than attempting to chase edible ghosts, like in the previous two experiments. The best performing agents would keep away from danger and consume as many pills as possible through each maze, this behaviour trait was so strong that even if a power pill was consumed, the agent would stay away from the edible ghosts. Due to this, the average scores achieved by these agents was slightly less than the others, however, the agents that were produced were more generalised and could perform better on mazes it hadn't seen before than agents evolved for the other experiments.

Overall, the agents produced by using GP significantly out performed their human counterpart (which achieved lower scores on all the mazes) and the agent produced from the unlimited-maze experiment demonstrates that the best general behaviour to the problem can be produced using GP.

\subsubsection{Training Camp}
In the real world, \emph{training camps} are used to teach a person basic skills which are required to perform a more complex task. An example of this is a football training camp where pupils are taught different aspects of playing football individually, such as controlling and passing the ball, shooting, dribbling, etc. They then use and apply these skills in a full game. Alhejali and Lucas extended their original GP approach with this idea of training camps\textsuperscript{\cite{trainingcamp}}.

A problem with the previous GP system\textsuperscript{\cite{EA1}} was that calculating the fitness of an individual based only on the score achieved by the agent can be misleading. You cannot deduce different behaviours or strategies employed simply by looking at the score or number of levels cleared (luck can often influence the scores, especially with a non-deterministic game like Ms. Pac-Man). This is where the training camp hopes to improve on the original design, it will produce scenarios which aim to produce and improve specific behavioural traits. These individual scenarios will make it much easier to produce test-cases and when an agent has been trained in several different situations, the authors hypothesise that it will perform better in a full game than their previous GP approach.

The function set used in this GP experiment\textsuperscript{\cite{trainingcamp}} is very similar to that in the previous system, except that the action terminals will now be produced using a training camp which will use GP to solve given scenarios to produce sub-agents. Once these sub-agents are evolved, they will be used as action terminals for a complete agent to play a full game. 

The scenarios used for this experiment have a goal which is directly related to the goals in a full game. A strategy will be produced to deal with each scenario. When producing these strategies, the GP system will only run while during the scenario, after which it will stop. The different scenarios are: clearing the maze, escaping the ghosts and chasing the ghosts. You can see how building advanced strategies for each of these scenarios should improve the agents performance during a full game.

The results of the experiment confirm the author's hypothesis, the training camp was shown to outperform the standard GP in terms of raw scores and stability. On average, the training camp scored around 1800 points more than the standard algorithm and it's maximum score was nearly 800 points greater. This means that the GP evolved using the training camp could apply behaviours learnt in the individual scenarios to great effect in the full game.

The use of training camps in the second generation of Alhejali and Lucas's experiment has clearly show an improvement on the first system. They conclude by commenting on the difficulty on producing viable test scenarios and how this system could be improved by designing a method where scenarios could be automatically generated, there were also concerns on the amount of CPU time needed to compute the different aspect of the training camp. Despite this, the research presented has shown that more successful agents can be produced by splitting the main problem into smaller, less complicated sub-problems which can be used to produce desirable behavioural traits for a final, complete agent.




\subsection{Agents}

\subsubsection{Ant Colony Optimisation}
The nature of \emph{Ant Colony Optimisation (ACO)} seems to make it a very good candidate solution to building a Ms. Pac-Man agent. ACO has traditionally been used in route planning applications, such as solving the \emph{Travelling Salesman Problem}\textsuperscript{\cite{ACOtech}}, moreover, the goals in Ms. Pac-Man are similar to that of route planning and can be quite easily applied to ACO. 

\emph{Pac-mAnt}\textsuperscript{\cite{pac-mant}} is an implementation of ACO applied to Ms. Pac-Man, it intends to present the promise of applying optimisation algorithms to video game agents. The authors explain how ACO can be easily applied to a problem which can be formalised by graphs which includes Ms. Pac-Man since the game map can be represented as a set of nodes. They then go on to explain how there are distinct differences between Ms. Pac-Man and a traditional optimisation problem, the first being that there is no clear destination throughout most of the game-play and secondly that the weights on each of the nodes will vary over time, due to the movement of the ghosts. These characteristics show that when an algorithm like ACO is to be applied to certain problems, the objectives will be different and the implementation will need to account for these.

Pac-mAnt's solution for solving the objectives of Ms. Pac-Man, while using an ACO based approach, is achieved by using two different types of ant, collectors and explorers. As usual with ACO, the ants will have a limited distance to move as not to impede on the limited CPU time available for each move Pac-Man must make. The collector ants role in this algorithm is to find the path with the greatest payout in terms of pills/power pills collected over a certain distance, they also take into account any edible ghosts through each route. In contrast to this, the explorer ants role is to find a path to safety for Pac-Man if there are ghosts nearby. At each time step in the game, ants of both types are launched down each different route Pac-Man could take, after the ants have moved the maximum distance allowed, the next move must be selected. In Pac-mAnt, this is a very simple decision, if a ghost is within a specified distance (a parameter set at the start of execution) Pac-Man will follow the explorer ant with the greatest amount of pheromone. If there is no ghost nearby, Pac-Man will follow the collector ant with the greatest amount of pheromone. This is a good solution to the problem explained above as the algorithm for Pac-mAnt now has the capability to work with two different objectives, collecting points and staying alive.

There are numerous different parameters which need to be optimised in Pac-mAnt's implementation of ACO including: max number of ants; min ghost distance; pheromone drop constants (different for each type of ant); pheromone evaporation constants (also different for each type); etc. Pac-mAnt attempts to find the globally optimised parameters by using a simple Genetic Algorithm. Each individual will contain a variable for each parameter and these are evolved using mutation and multipoint crossover. The fitness function measures the score of each game played with each individuals parameters. The population is decided by tournament selection.

The results from this research were promising, the Pac-mAnt agent could achieve a reasonably high score, although the authors comment on the noisy data available to the agent. This may have been caused by the screen-capture method of extracting data about the game's current state. This meant that in turn, the evaluation function of the parameter optimising Genetic Algorithm could not achieve the best results it could. However, despite this, I believe that further research could reveal a very capable and promising agent based on ACO.





\chapter{Cellular Automaton Based Approach}


\section{Introduction}
Most of the techniques used for video game AI are fairly old and traditional, approaches such as: Bayesian Programming, Heuristic Search, Planning, etc. are widely used to build AI components. The reason these techniques are so popular is because they have been applied to difficult problems which coexist in the real world and a virtual game world for decades. An example of such a problem is route planning, which can be solved using an appropriate Heuristic search algorithm. This style of game AI design has therefore left little room for developers to explore and experiment with new and novel techniques which may provide a way to produce 'smarter' AI players.

\subsection{Why use a Cellular Automaton?}
The aim for this project was to design, build, test and document an approach to video game AI which hasn't yet been attempted or popularised. As explained earlier, the game that will be used to test the approach will be Ms. Pac-Man and the technique that will applied to the AI player is a Cellular Automaton (CA).

There are many reasons why a CA was chosen as the basis for the Ms. Pac-Man controller. Firstly, there has been very little research activity in the multi-agent system area of Ms. Pac-Man AI or game AI as a whole. Secondly, we believe that the CA's holds a lot of promise in this area as they are very reactive to their environment and are very efficient to run. The also produce distinctive and complex behaviour which cannot be easily replicated using traditional algorithms.

Most of the previous approaches to Ms. Pac-Man controllers have focused on game intelligence and awareness from the Pac-Man's perspective, in our approach the Pac-Man makes very simple decisions based on it's local environment, we are essentially building a map for Pac-Man to follow rather than giving Pac-Man detailed rules for making decisions.

The name we have given our controller is \emph{CAp-man}.

\section{Design}

Our approach is inspired by traditional Cellular Automaton algorithms, it follows the general procedures and the way the grid of cells updates appears to follow the regular method of doing so. However, there are a few key differences between our algorithm and the traditional approach.

Firstly, the values of cells are represented by two different parts. The primary part is based on the different elements in the game of Pac-Man: Pills (\texttt{p}), Power-Pills (\texttt{P}), Ghosts (\texttt{G}), Edible Ghosts (\texttt{E}), Empty cells (\texttt{e}) and of course Pac-Man itself (\texttt{@}). As you can see we have denoted each element by a symbol. The secondary part of a cell's value is a decay index, this is a number which will specify how far a cell has propagated from it's source (more on this in the next section).

Normally in a CA, there is a set of rules which govern the way that the cells update their values. For example, in a simple system where cells can have one of two values, either \texttt{ON} or \texttt{OFF}, a rule could look like this:\linebreak
\centerline{\small{\texttt{FOREACH (cell.neighbour): IF (cell.neighbour.value ==  ON) $\rightarrow$ TURN ON}}}
This rule will iterate through a cell's neighbours and if any one of them have the value \texttt{ON}, the cell will change it's own value to \texttt{ON}. This is the general procedure for CA's, the update rules will change a cell's value based on it's neighbours values. Our approach however, operates differently to that - the cells will attempt to dominate and pass on their value to their neighbours. The way in which a cell will dominate depends on it's value. There are rules which control which cell values are more powerful (see section~\ref{subsec:htaw}). This method of updating cells results in a profoundly different effect to that of the tradition cell update, where a cells value will propagate from it's source as it dominates through its neighbours. Once the cell updates have been completed, Pac-Man will look at it's neighbouring cells and make a decision on which direction to travel, note that only the cells adjacent to Pac-Man will be considered. This means that Pac-Man has no knowledge of the game state apart from the values of it's direct neighbours. The decisions Pac-Man makes are based on very simple logic: move towards goals and/or away from danger. The power in this approach arises from the interactions from the CA as a whole, Pac-Man has limited knowledge of his environment, however the environment has total knowledge of the game state.

\begin{figure}[h!]
  \centering
    \includegraphics[width=1\textwidth]{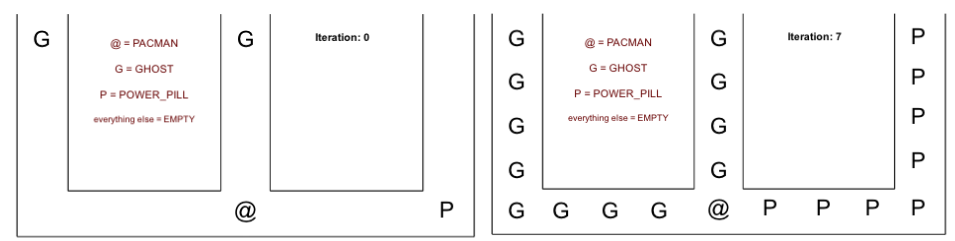}
  \caption{\small{An example showing the initial state CA and the updated state.}}
  \label{fig:f1}
\end{figure}

When the CA iterates through a specified number of updates, the result is essentially a map of game state. The powerful cells will dominate their neighbours and will propagate their value through the maze. See Figure~\ref{fig:f1}.

As you can see, the the left hand side shows the initial state and the right shows the final, updated state. The original sources have propagated out their values (all artefacts in the CA will dominate empty cells, apart from Pac-Man). At this point the state of the CA is stable, nothing will change given the current situation. Pac-Man must make a choice on which direction it should travel, in this case its an easy decision - he would move to the right, away from the danger (Ghosts) and towards a goal (Power-Pill).

There is however, an inherent problem with this approach. If there is a situation where each of Pac-Man's neighbour cells contain the same value, how can he make a decision on which one to follow? There needs to be another factor in the decision making process. This is where we introduce the notion of cell decay.

\begin{figure}[h!]
  \centering
    \includegraphics[width=1\textwidth]{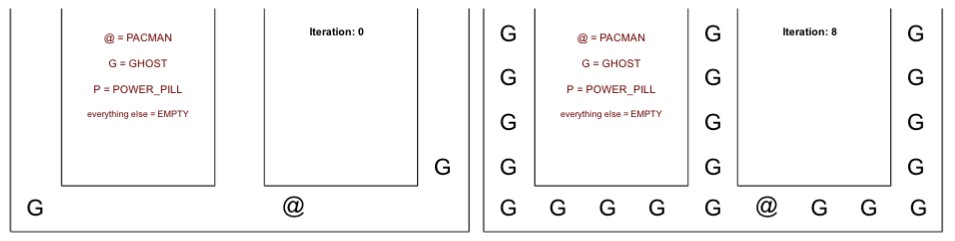}
  \caption{\small{An example showing a weakness with cell dominance.}}
  \label{fig:f2}
\end{figure}

\begin{figure}[h!]
  \centering
    \includegraphics[width=1\textwidth]{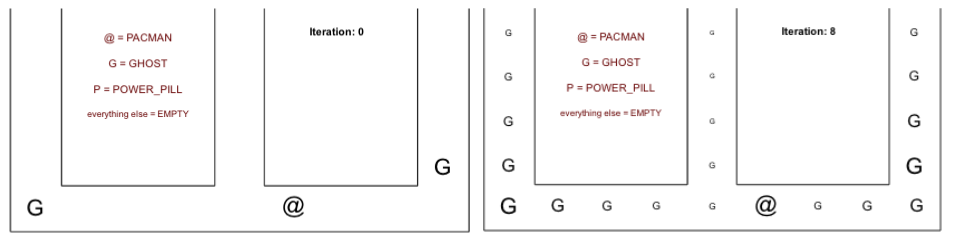}
  \caption{\small{An example showing the effect of cell decay.}}
  \label{fig:f3}
\end{figure}

You can see from the situation shown in Figure~\ref{fig:f2} that Pac-Man cannot make an informed decision when choosing his move (since he only takes his adjacent neighbours into account). We can see, as observers, that the moving left would be the correct decision, so Pac-Man moves away from the closer Ghost. By using cell decay we can make this information available to Pac-Man and still maintain the simple decision making process where the intelligence in the environment. In Figure~\ref{fig:f3} we observe that the dominated cells essentially become smaller as they propagate from the source and when Pac-Man comes to make his decision, he will know that the Ghost on his right hand side is closer, so will move away from it.

Essentially, if Pac-Man is presented with cells of the same value, he will choose the cell with the highest decay value (or lowest if the cells are Ghosts, since he is trying to move away from them).

This is a very effective and efficient solution to the problem described earlier. It allows for much more informed decisions, whilst not impeding the simple process and interactions between cells. The only problem with using cell decay is that we may encounter a situation where two sources are an equal distance apart (for one or more game steps). This will cause Pac-Man to swing back and fourth from the source cells (since the cell value and decay values are equal). To combat this, we will use a small amount of randomness to ensure that a route will always be chosen, however not too much that it effects the overall decay value, or the distance Pac-Man will assume that's between it and the source.

\subsection{Detailed Algorithm}
\label{subsec:htaw}

{
\raggedleft
\textbf{The different values a cell can take are:}
}

\begin{itemize}
\item Pac-Man $\rightarrow$ \texttt{@}
\item Pill $\rightarrow$ \texttt{p}
\item Power-Pill $\rightarrow$ \texttt{P}
\item Ghost $\rightarrow$ \texttt{G}
\item Edible Ghost $\rightarrow$ \texttt{E}
\item Empty $\rightarrow$ \texttt{e}
\end{itemize}

{
\raggedleft
\textbf{The domination rules for each cell value are:}
}

\begin{itemize}
\item \texttt{@ dominates []\footnote{Note that Pac-Man does not dominate any other cell}}
\item \texttt{E dominates [p, P, G, e]}
\item \texttt{p dominates [e]}
\item \texttt{P dominates [p, G, e]}
\item \texttt{G dominates [p, G, e]}
\item \texttt{e dominates []}
\end{itemize}

{
\raggedleft
\textbf{The best neighbour decision for Pac-Man (highest priority first, takes decay into account):}
}

\begin{enumerate}
\item \texttt{E}
\item \texttt{P}
\item \texttt{p}
\item \texttt{e}
\item \texttt{G}
\end{enumerate}

{
\raggedleft
\textbf{The decay rule (applies to every cell value except Pac-Man and empty):}
}
\[D = PD - \frac{PD}{10} + (R * \frac{PD}{1000})\]

{
\raggedleft
\textbf{where:}
}
	
		\begin{itemize}		
		\item \texttt{D = cell decay}
		\item \texttt{PD = previous cell decay}\footnote{Cells initial decay value = 1.}
		\item \texttt{R = random number ($-0.5 \leq 0 \leq 0.5$)}
		\end{itemize}

{
\raggedleft
\textbf{The CA algorithm:}
}

{
\raggedleft
\texttt{CA will be updated N times at each game step}
}

\begin{enumerate}
\item \texttt{WHILE (N != 0):}
\item \texttt{FOREACH NON-EMPTY cell:}
\item \texttt{DO cell.dominate(cell.neighbours)}
\item \texttt{END FOREACH}
\item \texttt{N = N - 1}
\item \texttt{END WHILE}
\item \texttt{RETURN PACMAN.bestMove(PACMAN.neighbours)}
\end{enumerate}

{
\raggedleft
\textbf{where:}
}
		\begin{itemize}
		\item \texttt{dominate(cell.neighbours)} is a function which will follow the domination rules according to \texttt{cell.value}. It will pass its cell.value onto any dominated cells.
		\item \texttt{bestMove(PACMAN.neighbours} is a function which will return the best direction Pac-Man can move in - based on it's neighbours.
		\end{itemize}

\section{Implementation}

Since the simulator contains all of the necessary code to run a game of Ms. Pac-Man, the only part that needed implementing on my behalf was the controller. The controller is implemented in Java (since this is what language the simulator is coded in) and contains the CA algorithm which was discussed in section~\ref{subsec:htaw}. The software was engineered using Agile Development methods.

\subsection{From Design to Code}

The grid to contain the cells for the CA is implemented as a 2-D array. The cells themselves are of type \texttt{Cell}, these objects contain all of the necessary data which is needed by the algorithm - cell value, decay value and information about the cell's location in the grid. The cell values are implemented into an \texttt{Enumeration} - since they and the domination rules are constant. Each cell value contains the domination rule associated with it, plus the decay rule. 

The simulator provides an abstract class which must be implemented in order to build a Ms. Pac-Man controller. This class contains one method which must be overridden - \texttt{getMove(Game game} where \texttt{game} is the current game state. This method is called by the game engine at every game step.

\subsection{Classes}

\begin{itemize}
\item \texttt{CApMan} - contains the CA grid, CA algorithm and other helper functions.
\item \texttt{CApManController} - implements the \texttt{Controller} abstract class provided by the simulator. This class will initialise the CA and run it on each game step.
\item \texttt{CApManDecay} - contains the decay function
\item \texttt{CApManConstants} - contains the cell values, domination rules and best-neighbour rule
\end{itemize}

\section{Implementation Testing}

The implementation of the CA controller was tested formally and informally. Unit testing was employed to ensure certain important, but trivial, functions produced the correct output (e.g. a function which will return the neighbours of any given cell, or a function which will update a cell with given parameters). Informal testing was used to monitor the behaviour of the CA in certain scenarios (e.g. moving away from an adjacent cell containing a ghost), this was done by checking a formatted output of the CA's state at the given time, essentially a large 2-D array. This was an important test as it gave us an easy way to determine if the controller reacted in the way we wanted to certain situations.



\section{Results}

Due to the noisy nature of non-deterministic video games, it is very difficult to yield good results for AI controllers which accurately represent their performance. This is especially true for Ms. Pac-Man as there is a very large amount of randomness in each game. This means that a 'good score' could be attributed to the luck of the player. To try to counteract this, each controller will play 100 full games against the same opponent. The average, minimum and maximum scores will be recorded along with the standard deviation for each controller. By playing this number of games the sprees of luck will hopefully be reduced and an accurate representation of the controllers performance will be achieved.

Alongside the scores for each controller, there will also be an evaluation of their behaviour. This will allow for a more in-depth analysis of the performance of each controller and will allow us to isolate desired behavioural traits which the controllers exhibit.

In the first stage of the evaluation for \emph{CAp-man} we need to find what the optimum number of updates for each situation is. To do this in a time efficient manner, a two-stage evaluation was conducted. In the first stage, broad ranges of updates were used to try to discover what the optimum range was. The second stage was then used to attempt to find the best controller in that optimum range.

Each different controller will play the same number of games against the same opponent, which is a set of random ghosts. This ghost controller will cause the ghosts to choose a random direction to travel each time they arrive at a junction in the maze (including the opposite direction they were moving in). The reason this controller will provide a good test is because it is completely non-deterministic, there is no way of predicting what state the ghosts will be in at any point throughout the game. The non-determinism of the controller will be the source of unexpected behaviour each controller will have to deal with, which will be challenging.

\subsection{Comparison with Baseline Controllers}
Our CA based controller will be compared to two other controllers to determine its performance. The main factor which will be compared between the controllers will be the score achieved by each of them. However, the actual behaviour of the controllers will also be considered because we want to evaluate the different behavioural traits and decide which (applicable) controller shows more promise for future development.

The first controller we will use for the comparison is a simple rule based controller, called \emph{StarterPacMan} (which is included as part of the simulator package). This controller follows a set of static, hand-coded rules in decreasing priority:
\texttt{
\begin{enumerate}
\item
If any non-edible Ghost is within close proximity (20 nodes), run away.
\item
Chase nearest edible Ghost  (if any exist).
\item
Go to nearest Pill/Power Pill
\end{enumerate}
}

The second controller to be used is a human based controller. There will be 10 participants, each of different skill levels (from beginners to advanced), and each of them will play 10 games against the ghost controller. The behaviour of each human player will also be examined.

\subsection{Controller Performance}

In this section, there are graphs to show how the performance of CAp-Man changes as the number of updates is modified. It will show the initial evaluation of the controller (using large differences between number of updates to find the optimum range) and the second, more in depth evaluation (testing every number of updates within the discovered optimum range).

Figure~\ref{fig:t1} shows the results from the first evaluation of CAp-Man. In this experiment the differences between the number of CA updates it quite large. We are trying to find in which range the global optimum number of updates lies.

\begin{figure}[h!]
\begin{center}
    \begin{tabular}{| c | c | c | c | c |}
    \hline

	Update Steps & Average & Standard Deviation & Min Score & Max Score \\ \hline   
	
	1 & 4547 & 2656 & 1380 & 13010 \\ 
	11 & 49480 & 12944 & 14150 & 62680 \\ 
	21 & 49030 & 12269 & 6900 & 61440 \\ 
	31 & 43492 & 11332 & 14050 & 59250 \\ 
	41 & 35395 & 11855 & 8780 & 50350 \\ 
	51 & 28794 & 8605 & 4920 & 44160 \\ 
	61 & 26193 & 8376 & 6472 & 39014 \\ 
	71 & 21039 & 9028 & 2049 & 30951 \\ 
	81 & 18038 & 7049 & 3394 & 23059 \\ 
	91 & 16883 & 6785 & 2903 & 19038 \\ 
    
    \hline
    \end{tabular}
      \caption{\small{Table showing data from 1st evaluation of CAp-Man performance.}}
  \label{fig:t1}
\end{center}
\end{figure}

Figure~\ref{fig:t1} shows the results for the second evaluation. Here we look into more detail within the optimum range - between 11 and 31.

\begin{figure}[h!]
\begin{center}
    \begin{tabular}{| c | c | c | c | c |}
    \hline

	Update Steps & Average & Standard Deviation & Min Score & Max Score \\ \hline   
	
	11 & 49480 & 12944 & 14150 & 62680 \\ 
	12 & 49913 & 12323 & 9030 & 63780 \\ 
	13 & 47892 & 11083 & 13750 & 60290 \\ 
	14 & 52894 & 11059 & 10930 & 65060 \\ 
	15 & 54638 & 13519 & 9400 & 63010 \\ 
	16 & 56724 & 11079 & 15730 & 64790 \\ 
	17 & 57829 & 12031 & 10220 & 67880 \\ 
	18 & 57001 & 10993 & 13380 & 63450 \\ 
	19 & 53949 & 13772 & 8040 & 58180 \\ 
	20 & 52890 & 11763 & 10380 & 56290 \\ 
	21 & 49030 & 12269 & 6900 & 61440 \\ 
	22 & 52845 & 13092 & 7810 & 57280 \\ 
	23 & 49152 & 11038 & 5920 & 60390 \\ 
	24 & 50032 & 10472 & 8990 & 54190 \\ 
	25 & 48721 & 9013 & 13750 & 52100 \\ 
	26 & 48109 & 8242 & 4820 & 53490 \\ 
	27 & 47298 & 10484 & 7890 & 50210 \\ 
	28 & 46723 & 9623 & 6520 & 54660 \\ 
	29 & 44672 & 11852 & 7880 & 55220 \\ 
	30 & 41274 & 9624 & 11250 & 55260 \\ 
	31 & 43492 & 11332 & 14050 & 53250 \\
    
    \hline
    \end{tabular}
      \caption{\small{Table showing data from 2nd evaluation of CAp-Man performance.}}
  \label{fig:t2}
\end{center}
\end{figure}

\begin{figure}[h!]
  \centering
    \includegraphics[width=1\textwidth]{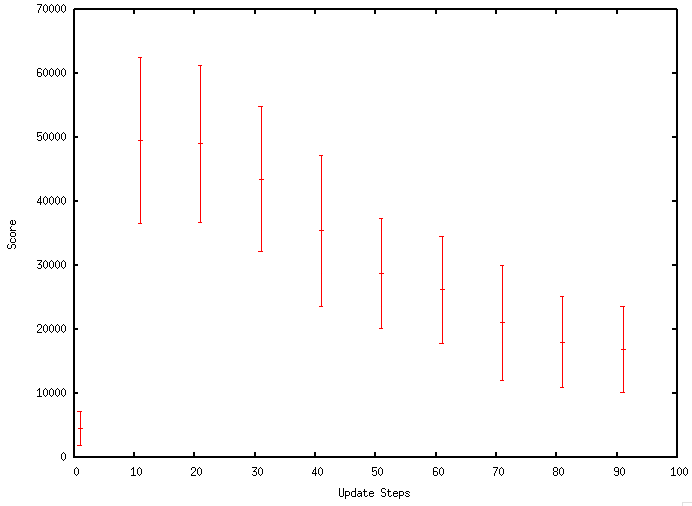}
  \caption{\small{Graph showing experiments from 1st evaluation of CAp-Man performance.}}
  \label{fig:g1}
\end{figure}

\begin{figure}[h!]
  \centering
    \includegraphics[width=1\textwidth]{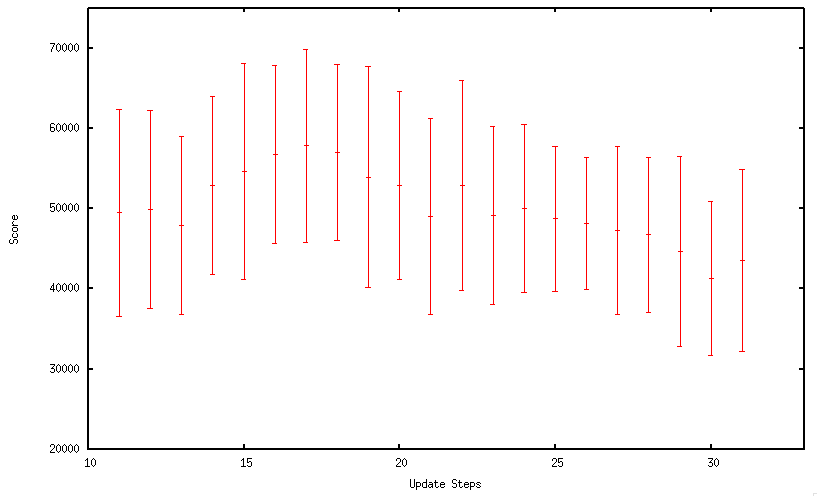}
  \caption{\small{Graph showing experiments from 2nd evaluation of CAp-Man performance.}}
  \label{fig:g2}
\end{figure}

You can see how the scores achieved by the controller increase by a large amount initially (see Figure~\ref{fig:g1}), this is because in the first test the CA only performed a single update - which makes the controller very greedy and will run into danger. The performance increases dramatically when the number of updates is increased, but starts to decrease as this number gets larger - this is due to a sort of confusion effect. When the number of updates is very large, information about the game state, which may have no relevance to Pac-Man at his current position, starts to warp the map of the environment created by the CA. Goals or dangers which are a large distance away from Pac-Man will influence his decisions which causes naivety to his local surroundings (which may contain a closer goal). The highest performing CAp-Man controller used 17 updates as you can see from the graph (Figure~\ref{fig:g2}).

The scores achieved from the 10 human participants are shown in Figure~\ref{fig:t3}

\begin{figure}[h!]
\begin{center}
    \begin{tabular}{| c | c | c | c | c |}
    \hline

	Human & Average & Standard Deviation & Min Score & Max Score \\ \hline   
	
	1 & 23065 & 5920 & 14029 & 31848 \\ 
	2 & 12942 & 3235 & 2064 & 14762 \\ 
	3 & 41028 & 9048 & 19050 &  45034\\ 
	4 & 32019 & 3854 & 17284 & 39174 \\ 
	5 & 5968 & 1937 & 4527 & 8376 \\ 
	6 & 19003 & 4632 & 11048 & 23852 \\ 
	7 & 43621 & 2190 & 34858 & 46382 \\ 
	8 & 21843 & 8028 & 12848 & 45673 \\ 
	9 & 9083 & 3624 & 5319 & 15099 \\ 
	10 & 13980 & 5831 & 5436 & 32953 \\ 
    
    \hline
    \end{tabular}
      \caption{\small{Table showing data from human-based controllers.}}
  \label{fig:t3}
\end{center}
\end{figure}

\begin{figure}[h!]
\begin{center}
    \begin{tabular}{| c | c | c | c | c |}
    \hline

	Controller & Average & Standard Deviation & Min Score & Max Score \\ \hline 
	CAp-Man - 17 & 57829 & 12031 & 10220 & 67880 \\  
	StarterPacMan & 34270 & 3049 & 30490 & 35900 \\ 
	Human (best) & 43621 & 2190 & 34858 & 46382 \\ 

    \hline
    \end{tabular}
      \caption{\small{Table showing data for all of the Ms. Pac-Man controllers used.}}
  \label{fig:t4}
\end{center}
\end{figure} 

\subsection{Controller Behaviour}

CAp-Man showed a lot of varied and interesting behaviour for each different number of updates. At each of the extremes (very low or very high numbers of updates) the controller seemed to act very greedily. It would pursue goals that were far beyond it's reach and would put itself into unnecessary danger. The best performing CAp-Man controller was found towards the middle of the extreme ranges of updates. This controller seemed to balance risk taking and safety quite well and was incredibly good at getting out of danger in tight situations. It showed very reactive behaviour and will quickly change it's direction to get out of danger. It completes many levels throughout a game and seems to prioritise the collection of Pills. Overall this controller played the game in a safe manner and would not pursue goals if a ghost was nearby. This is what allowed it to achieve the highest scores - the time it would stay alive in the game was the longest out of all the other Ms. Pac-Man controllers. It did however show a degree of naivety and will sometimes put itself into inescapable situations and would thus loose a life. The reason for this is due to the fact that the controller doesn't consider the dangerous parts of the maze. For example, if there is part of a maze which has only two entrances, the controller will happily guide Pac-Man into it if there is a goal. If Ghosts then proceed to move through both entrances, Pac-Man is stuck and will loose a life. CAp-Man does not have any awareness of the maze as an entity, it simply uses it as a base for the CA. If the algorithm was altered to take these dangerous situations into account, the problem could be avoided.

The StarterPacMan showed very different behaviour to that of CAp-Man. It plays the same way in almost every game and almost mirrors its route each time. It seemed to be a very greedy controller and would pursue any edible Ghost in the game. It's Ghost avoidance strategy is quite weak as it bases it's decision purely on the distances between it and the Ghosts. This means that it looses its lives quickly and the time spent in the game is reduced. It does however show that even a greedy controller, which looses fairly quickly, can achieve a reasonable score. This is because of it's greed and the fact it chases edible Ghosts. With some tweaking of it's rules, the performance of this controller could be improved.


\chapter{Conclusion}

\section{A Successful Approach?}

Our CA based Ms. Pac-Man controller has proved to be reactive and effective. It has achieved very high scores which surpass the simple Pac-Man controller and the human based controller. Its behaviour is very interesting and shows many desired traits - like the way it escapes from the Ghosts. Moreover, it shows promise for future research and development.

However, it does have some weaknesses. It can be very naive when coming into contact with the ghosts and can place itself into inescapable situations. This problem shows that more work does need to be done, but I believe that with extra functionality and optimisation, the performance of CAp-Man could be improved hugely.

\section{Contributions to the Field}

This project has shown that the application of untraditional and novel techniques to game AI has definite promise for future research. It shows that using games like Ms. Pac-Man as test beds for AI techniques can be a very effective way of determining their performance and analysing their behaviour. Furthermore, games can be used as part of a benchmarking procedure for various AI techniques.

\section{Evaluation of the Mini Project}

I believe that this project was a success, I proposed questions at the start and have answered them throughout this report. Given more time, I would like to have produced several different variations of the CA based controller, with different rules and complexities, and compared them. This would give a deeper understanding into the way CA's could be used in game AI and could show more reasons to pursue further research in this context.

\chapter{Future Work}

This project is open to a large amount of potential future research. Since the model we used in this particular approach used simple rules and interactions, there are many different ways it could be extended and potentially improved.

\section{Extension 1}

The number of updates performed on the CA is static throughout the game. We ran experiments to try and locate the global optimum for this. There is however an uncertainty in whether or not there is an optimum for this parameter since we don't know the effect of using different numbers of updates midway through a game. A potential improvement to the static update parameter could be to introduce a dynamic one. The number of CA updates performed at each game step could be altered to improve efficiency and effectiveness. For example, if we reach a stable state in the CA, there is no point in continuing to try and update the cells. On the other hand if we iterate up to our update number, we have to stop and get the best move we can from the current state of the CA, what if there was a better move which could be realised if we had performed more updates. This is where a dynamic update parameter which can evolve on the fly could be very useful and increase the performance of our approach.

\section{Extension 2}

Another potential improvement which could be made is to build more complicated rules for the CA to operate with. The rules we used on CAp-Man are very simple , yet they produce very complicated behaviour. If we increase the complexity of our rules - by adding extra functionality to make the cell domination more effective, we could increase the global awareness of the CA and potentially increase the performance of the controller. For instance, we could use neural networks which would take a cell, it's neighbour and part of the game state as an input and return a boolean value to determine if a cell should dominate. The rules could also be dynamic, perhaps evolving over time to improve their effectiveness. This would be a good improvement if the controller were to face many different types of Ghost opponents over the course of a game.

\section{Extension 3}

The final extension is state look ahead. This would try to minimise the number of bad/wrong moves made by Pac-Man by exploring into potential future states which would arise if Pac-Man makes a certain move. The future states would be analysed for undesired outcomes which could reduce the performance. It could also be used to verify if a potential move is good or not. This would of course be very computationally expensive - as it means expanding into many different game states (you would have to consider the moves each Ghost could make as well), but could prove to be a very good improvement to the effectiveness of the controller.


\nocite{*}
\bibliographystyle{plain}
\bibliography{research}

\clearpage
\addcontentsline{toc}{chapter}{Bibliography}

\begin{thebibliography}{1}

  \bibitem{pman}
  \emph{Pac-Man}.
  \url{http://en.wikipedia.org/wiki/Pac-Man}, Wikipedia.
  
    \bibitem{mspman}
  \emph{Ms. Pac-Man}.
  \url{http://en.wikipedia.org/wiki/Ms._Pac-Man}, Wikipedia.
  
  \bibitem{mario}
  \emph{Mario AI}.
  \url{http://www.marioai.org/}, Julian Togelius.
  
  \bibitem{torcs}
  \emph{TORCS - The Open Racing Car Simulator}.
  \url{http://torcs.sourceforge.net/index.php}, Bernhard Wymann.
  
  \bibitem{unre}
  \emph{2K BotPrize}.
  \url{http://botprize.org/}, Philip Hingston.
  
  \bibitem{star}
  \emph{StarCraft AIIDE Competition}.
  \url{http://webdocs.cs.ualberta.ca/~cdavid/starcraftaicomp/}, Michael Buro, David Churchill.
  
  \bibitem{pvgl}
  \emph{Ms. Pac-Man vs Ghosts League}.
  \url{http://www.pacman-vs-ghosts.net/}, Game Intelligence Group, University of Essex.
  
    \bibitem{wolf}
  \emph{Cellular Automaton}.
  \url{http://mathworld.wolfram.com/CellularAutomaton.html}, Wolfram Mathworld.

  \bibitem{step}
 Wolfram, S;
  \emph{Statistical mechanics of cellular automata}. reviewers of Modern Physics, Vol 55 No. 3 Pg: 601 - 644,
  1983, The American Physical Society, USA.

  \bibitem{GP1}
  Cramer, N A;
  \emph{A Representation for the adaptive generation of simple sequential programs}. Proc. of an Intl. Conf. on Genetic Algorithms and their Applications, Pg: 183 - 187,
  1985, Carnegie-Mellon University, Pittsburgh, PA.
  
  \bibitem{GP2}
  Koza, J;
  \emph{Genetic Programming: On the Programming of Computers by Means of Natural Selection}.
  1992, Cambridge, MA, The MIT Press.

  
  
  \bibitem{pam2}
  Thawonmas, Ruck; Matsumoto, Hiroshi;
  \emph{Automatic Controller of Ms. Pac-Man and Its Performance: Winner of the IEEE CEC 2009 Software Agent Ms. Pac-Man Competition}.
  Graduate School of Science and Engineering, Ritsumeikan University, Japan, 2009.
  
    \bibitem{ACOtech}
  Dorigo, Marco K; Strutzel, Thomas;
  \emph{Ant Colony Optimization: Overview and Recent Advances}. IRIDIA – Technical Report Series,  Number: TR/IRIDIA/2009-013, 2009, IRIDIA.
  
  
    \bibitem{pac-mant}
  Emilio, Martin; Moises, Martinez; Gustavo, Recio; Yago, Maez;
  \emph{Pac-mAnt: Optimization Based on Ant Colonies Applied to Developing an Agent for Ms. Pac-Man}. IEEE Conference on Computational Intelligence and Games, Pg: 458 - 464,
  2010, IEEE.
  
  \bibitem{gramma}
  Galvan-Lopex, Edgar; Swafford, John M; O'Neill, Micael; Brabazon, Anthony;
  \emph{Evolving a Ms. Pac-Man Agent Controller Using Grammatical Evolution}. Applications of Evolutionary Computation, EvoApplications, Istanbul, Turkey, 2010, Springer.
  
  \bibitem{pam3}
  Thawonmas, Ruck; Ashida, Takashi;
  \emph{Evolution Strategy for Optimizing Parameters in Ms. Pac-Man Controller ICE Pambush 3}. IEEE Conference on Computational Intelligence and Games,  Pg: 235 - 240, 2010, IEEE.
  
       \bibitem{EA1}
  Alhejali, Atif, M; Lucas, Simon, M;
  \emph{Evolving Diverse Ms. Pac-Man Playing Agents Using Genetic Programming}.
  Game Intelligence Group, School of Computer Science and Electronic Engineering, University of Essex, UK.

\bibitem{trainingcamp}
  Alhejali, Atif M; Lucas, Simon M;
  \emph{Using a Training Camp with Genetic Programming to Evolve Ms Pac-Man Agents}. IEEE Conference on Computational Intelligence and Games,  Pg: 118 - 125, 2011, IEEE.
  
  
  
  \bibitem{inforet}
  Brandstetter, Matthias F; Ahmadi, Samad;
  \emph{Reactive Control of Ms. Pac-Man using Information Retrieval based on Genetic Programming}. IEEE Conference on Computational Intelligence and Games,  Pg: 250 - 256, 2012, IEEE.
  
  
  

\end{thebibliography}

\end{document}